\documentclass[11pt]{article}
\usepackage[hyperref]{ccl2023-en}
\usepackage{times}
\usepackage{url}
\usepackage{latexsym}
\usepackage{fancyhdr}
\usepackage{algorithm}  
\usepackage{algorithmic}
\usepackage{amsmath}
\usepackage{amsfonts}
\usepackage{graphicx}
\usepackage{multirow}
\usepackage{subcaption}

\title{Rethinking Label Smoothing on Multi-hop Question Answering}
\author{
Zhangyue Yin\textsuperscript{$\diamondsuit$\thanks{{} {} Equal contribution. }}\quad
Yuxin Wang\textsuperscript{$\diamondsuit$}\footnotemark[1] \quad
Xiannian Hu\textsuperscript{$\diamondsuit$} \quad
Yiguang Wu\textsuperscript{$\diamondsuit$} \quad
Hang Yan\textsuperscript{$\diamondsuit$} \\
\bf{
Xinyu Zhang\textsuperscript{$\spadesuit$} \quad
Zhao Cao\textsuperscript{$\spadesuit$} \quad
Xuanjing Huang\textsuperscript{$\diamondsuit$} \quad
Xipeng Qiu\textsuperscript{$\diamondsuit$}\thanks{\ \ \ Corresponding author.}
}\\
\textsuperscript{$\diamondsuit$}School of Computer Science, Fudan University \\ 
\textsuperscript{$\spadesuit$}Huawei Poisson Lab \\
\texttt{\{yinzy21,wangyuxin21,xnhu21\}@m.fudan.edu.cn} \\
\texttt{\{ygwu20,hyan19,xjhuang,xpqiu\}@fudan.edu.cn} \\
\texttt{\{zhangxinyu35,caozhao1\}@huawei.com}
}
\date{}

\begin{document}
\maketitle
\begin{abstract}
Multi-Hop Question Answering (MHQA) is a significant area in question answering, requiring multiple reasoning components, including document retrieval, supporting sentence prediction, and answer span extraction. 
In this work, we analyze the primary factors limiting the performance of multi-hop reasoning and introduce label smoothing into the MHQA task. 
This is aimed at enhancing the generalization capabilities of MHQA systems and mitigating overfitting of answer spans and reasoning paths in training set. 
We propose a novel label smoothing technique, F1 Smoothing, which incorporates uncertainty into the learning process and is specifically tailored for Machine Reading Comprehension (MRC) tasks. 
Inspired by the principles of curriculum learning, we introduce the Linear Decay Label Smoothing Algorithm (LDLA), which progressively reduces uncertainty throughout the training process.
Experiment on the HotpotQA dataset demonstrates the effectiveness of our methods in enhancing performance and generalizability in multi-hop reasoning, achieving new state-of-the-art results on the leaderboard.

\end{abstract}

\section{Introduction}
\label{sec:intro}

\cclfootnote{
    %
    %
    \hspace{-0.65cm}  
    \textcopyright 2023 China National Conference on Computational Linguistics

    \noindent Published under Creative Commons Attribution 4.0 International License
}

Multi-Hop Question Answering (MHQA) is a rapidly evolving research area within question answering that involves answering complex questions by gathering information from multiple sources~\cite{Asai2020Learning,chen2021multihop}. This requires a model capable of performing several reasoning steps and handling diverse information sources~\cite{mavi2022survey}. In recent years, MHQA has attracted significant interest from researchers due to its potential for addressing real-world problems. The mainstream approach to MHQA typically incorporates several components, including a document retriever, a supporting sentence selector, and a reading comprehension module~\cite{tu2020select,li2022easy}. These components collaborate to accurately retrieve and integrate relevant information from multiple sources, ultimately providing a precise answer to the given question~\cite{feldman-el-yaniv-2019-multi}.

\begin{figure*}[t]
    \centering
    \begin{minipage}{0.48\textwidth}
    \centering
    \includegraphics[width=1\linewidth]{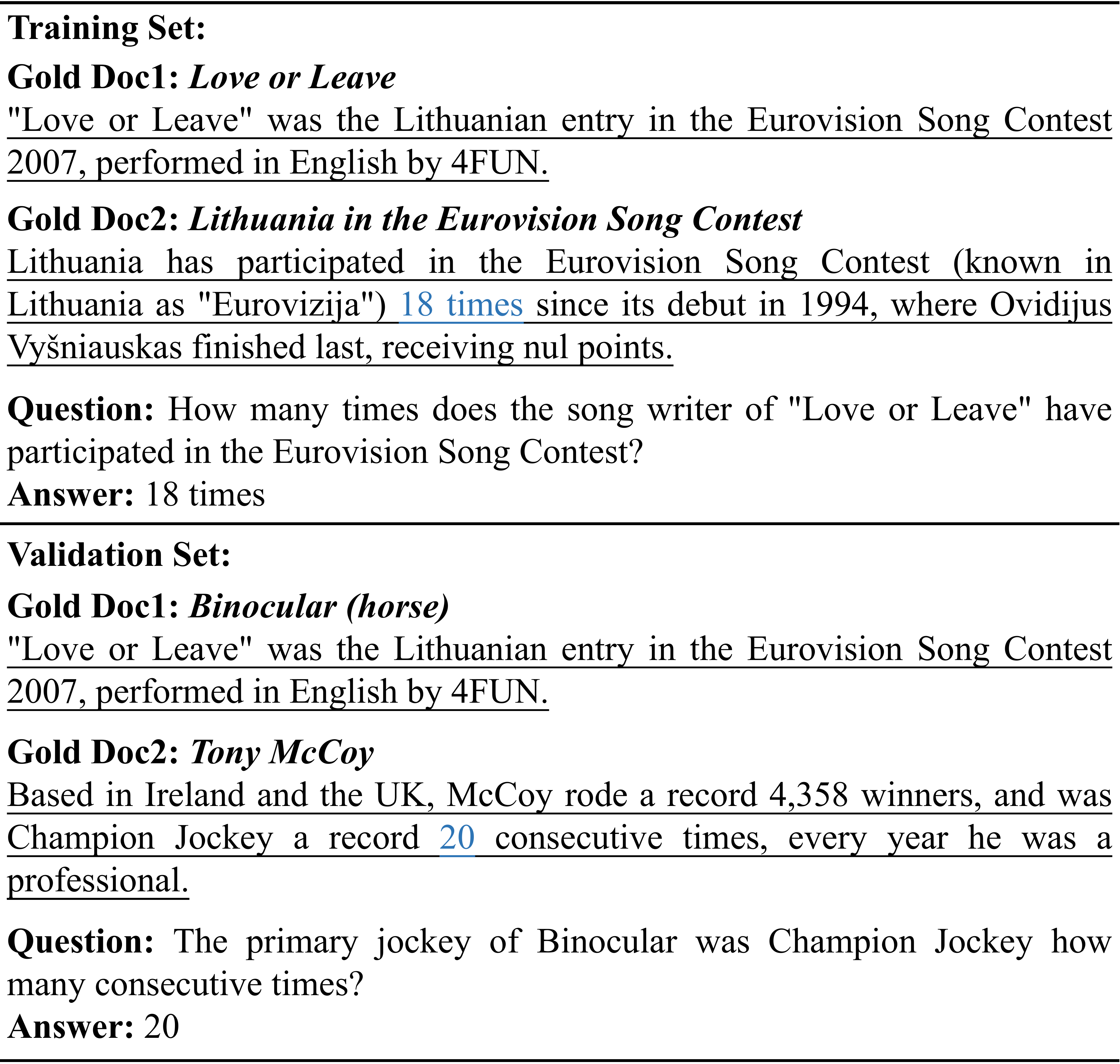}
    \vspace{-1.5em}
    \caption{Different Answer Span.}
    \label{fig:intro-answer-span}
    \end{minipage}
    \hfill
    \begin{minipage}{0.48\textwidth}
    \centering
    \includegraphics[width=1\linewidth]{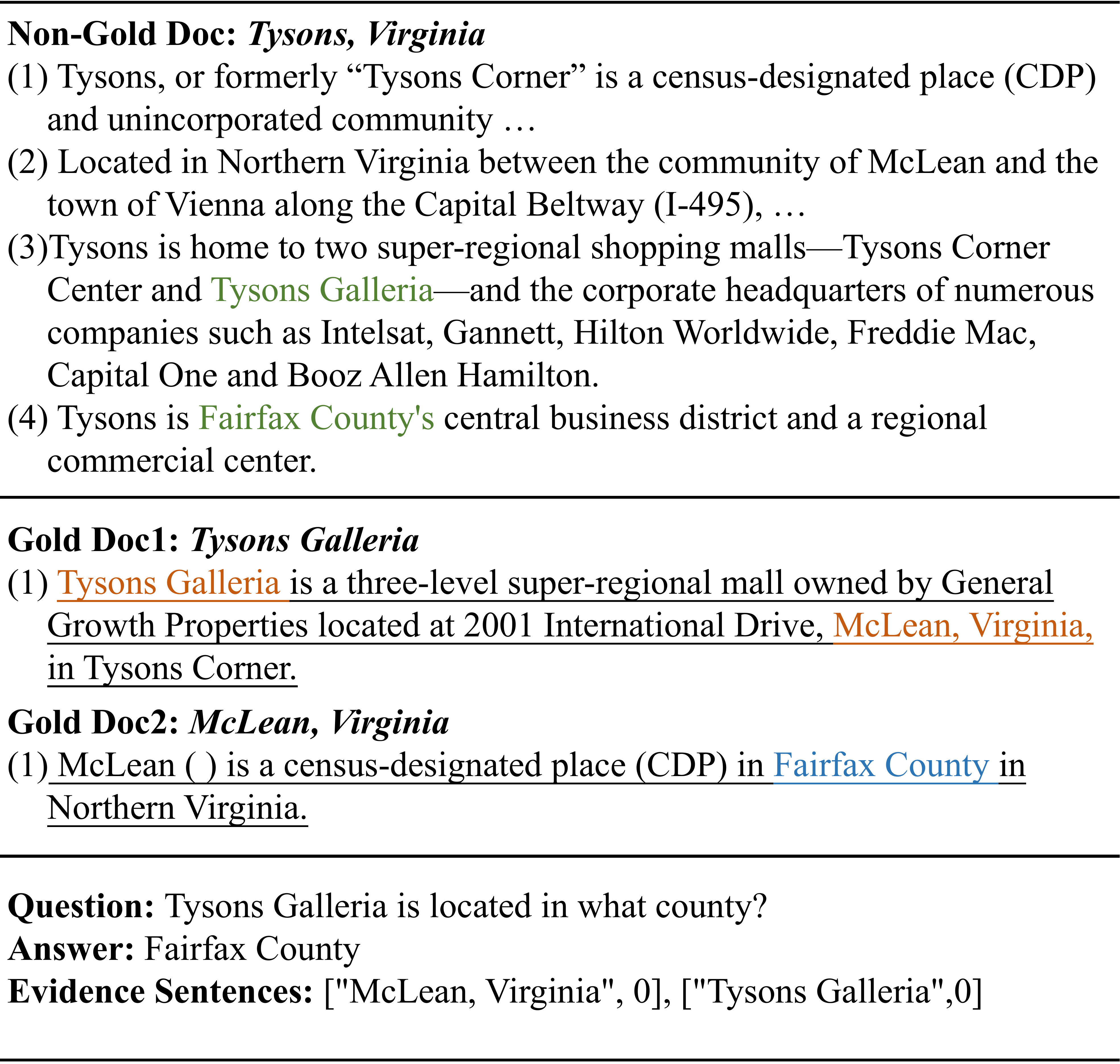}
    \vspace{-1.5em}
    \caption{Multiple Feasible Reasoning Paths.}
    \label{fig:intro-reasoning-paths}
    \end{minipage}
    \vspace{-.5em}
    \caption{Causes of errors in answer span and multi-hop reasoning in the HotpotQA dataset~\cite{Yang2018HotpotQAAD}. In Figure~\ref{fig:intro-answer-span}, the answer from the training set contains a quantifier, while the answer from the validation set does not. In Figure~\ref{fig:intro-reasoning-paths}, the correct answer can be inferred using a non-gold document without requiring information from gold document.}
    \label{fig:intro}
    \vspace{-1em}
\end{figure*}

MHQA models have shown remarkable capabilities in multi-hop reasoning. However, they still struggle with answer span errors and multi-hop reasoning errors. A recent study by S2G~\cite{wu2021graph} reveals that the primary error source is answer span errors, constituting 74.55\%, followed by multi-hop reasoning errors. We identify that answer span errors arise from differences in the annotation of answer spans between the training and validation sets. As depicted in Figure~\ref{fig:intro-answer-span}, the training set answer includes the quantifier ``times'', which is notably missing in the validation set. We observe that such discrepancies in answer spans are prevalent across both the training and validation sets. This demands that models possess a robust ability to generalize answer spans, thereby preventing overfitting to a specific answer span distribution present in the training set.

Furthermore, we discover the existence of unannotated yet feasible multi-hop reasoning paths within the training set. As depicted in Figure~\ref{fig:intro-reasoning-paths}, a non-gold document ``Tysons, Virginia'' contains essential information to deduce the answer ``Fairfax County'', but is marked as irrelevant. During training, this forces the model to ignore such reasoning paths and solely rely on the annotated ones. This potentially leads models to overfit specific multi-hop reasoning patterns labeled in the training set, consequently impairing their generalization capabilities on test sets. Hence, these observations naturally lead us to the research question we explore in this paper: \textit{How can we prevent MHQA models from overfitting answer spans and reasoning paths in the training set?}

Label smoothing has proven to be a highly effective technique for mitigating overfitting~\cite{Mller2019WhenDL,lukasik2020does,xu2020towards}, and it has been extensively employed across a diverse range of machine learning researches~\cite{szegedy2016rethinking,yuan2020revisiting,li2020regularization}. In this study, we pioneer the application of label smoothing to multi-hop reasoning tasks, aiming to reduce overfitting of answer spans and reasoning paths. Our proposed MHQA model, termed $\mathbf{R}^3$, integrates three key components: \textbf{R}etrieval, \textbf{R}efinement, and \textbf{R}eading Comprehension. 

Inspired by the F1 score, a widely used metric for evaluating Machine Reading Comprehension (MRC) task performance, we develop F1 Smoothing, a novel technique that calculates the importance of each token within the smooth distribution. Moreover, we incorporate curriculum learning~\cite{bengio2009curriculum} and devise the \textbf{L}inear \textbf{D}ecay \textbf{L}abel Smoothing \textbf{A}lgorithm (LDLA), which gradually reduces the smoothing weight, allowing the model to focus on more challenging samples during training. Experimental results on the HotpotQA dataset~\cite{Yang2018HotpotQAAD} demonstrate that incorporating F1 smoothing and LDLA into the $\mathbf{R}^3$ model significantly enhances performance in document retrieval, supporting sentence prediction, and answer span extraction, achieving new state-of-the-art results among all published works. Our main contributions are as follows:

\begin{itemize}
\item To our best knowledge, we are the first to adapt label smoothing for multi-hop reasoning tasks, encapsulated within our innovative $\mathbf{R}^3$ framework, featuring retrieval, refinement, and reading comprehension modules.
\item We propose F1 smoothing, a pioneering label smoothing method tailored for MRC tasks, which alleviates errors caused by answer span discrepancies.
\item We present the Linear Decay Label Smoothing Algorithm (LDLA), an innovative approach that combines the principles of curriculum learning for progressive training.
\item Experiment on the HotpotQA dataset demonstrates that label smoothing effectively enhances the MHQA model's performance, achieving new state-of-the-art performance on the leaderborad.
\end{itemize}

\section{Related Work}
\label{sec:related_work}

\vspace{-.1em}
\paragraph{Label Smoothing.} Label smoothing is a regularization technique first introduced in computer vision to improve classification accuracy on ImageNet~\cite{szegedy2016rethinking}. The basic idea of label smoothing is to soften the distribution of true labels by replacing their one-hot encoding with a smoother distribution. This approach encourages the model to be less confident in its predictions and consider a broader range of possibilities, reducing overfitting and enhancing generalization~\cite{pereyra2017regularizing,Mller2019WhenDL,lukasik2020does}. Label smoothing has been widely adopted across various natural language processing tasks, including speech recognition~\cite{Chorowski2017TowardsBD}, document retrieval~\cite{penha2021weakly}, dialogue generation~\cite{saha2021similarity}, and neural machine translation~\cite{gao2020towards,lukasik2020semantic,gracca2019generalizing}.

Recent studies are increasingly concentrating on enhancing and refining the conventional methods of label smoothing. For example, \newcite{xu2020towards} suggest the \textbf{T}wo-\textbf{S}tage \textbf{LA}bel (TSLA) smoothing algorithm, which employs a smoothing distribution in the first stage and the original distribution in the second stage. Experimental results demonstrate that TSLA effectively promotes training convergence and enhances performance. \newcite{penha2021weakly} introduce label smoothing for retrieval tasks and propose using BM25 to compute the label smoothing distribution, which outperforms the uniform distribution. \newcite{zhao2020robust} propose Word Overlapping, which uses maximum likelihood estimation~\cite{su2020label} to optimize the target distribution during training.

\vspace{-.1em}
\paragraph{Multi-hop Question Answering.} Multi-hop reading comprehension (MHRC) is a challenging task in the field of machine reading comprehension (MRC) that closely resembles the human reasoning process in real-world scenarios. Consequently, it has gained significant attention in the field of natural language understanding in recent years. Several datasets have been developed to foster research in this area, including HotpotQA~\cite{Yang2018HotpotQAAD}, WikiHop~\cite{welbl2018constructing}, and NarrativeQA~\cite{kovcisky2018narrativeqa}. Among these, HotpotQA~\cite{Yang2018HotpotQAAD} is particularly representative and challenging, as it requires the model to not only extract the correct answer span from the context but also identify a series of supporting sentences as evidence for MHRC.

Recent advances in MHRC have led to the development of several graph-free models, such as QUARK~\cite{groeneveld2020simple}, C2FReader~\cite{shao2020graph}, and S2G~\cite{wu2021graph}, which have challenged the dominance of previous graph-based approaches like DFGN~\cite{qiu2019dynamically}, SAE~\cite{tu2020select}, and HGN~\cite{fang2019hierarchical}. C2FReader~\cite{shao2020graph} suggests that the performance difference between graph attention and self-attention is minimal, while S2G's~\cite{wu2021graph} strong performance demonstrates the potential of graph-free modeling in MHRC. FE2H~\cite{li2022easy}, which uses a two-stage selector and a multi-task reader, significantly enhances the performance on HotpotQA, indicating that pre-trained language models are sufficient for modeling multi-hop reasoning. However, these approaches still suffer from answer spanning errors and multi-hop reasoning errors, primarily attributable to their restricted generalization abilities in multi-hop reasoning.

\section{Framework}
\label{sec:framework}
\begin{figure*}[t]
\centering
\includegraphics[width=1\linewidth]{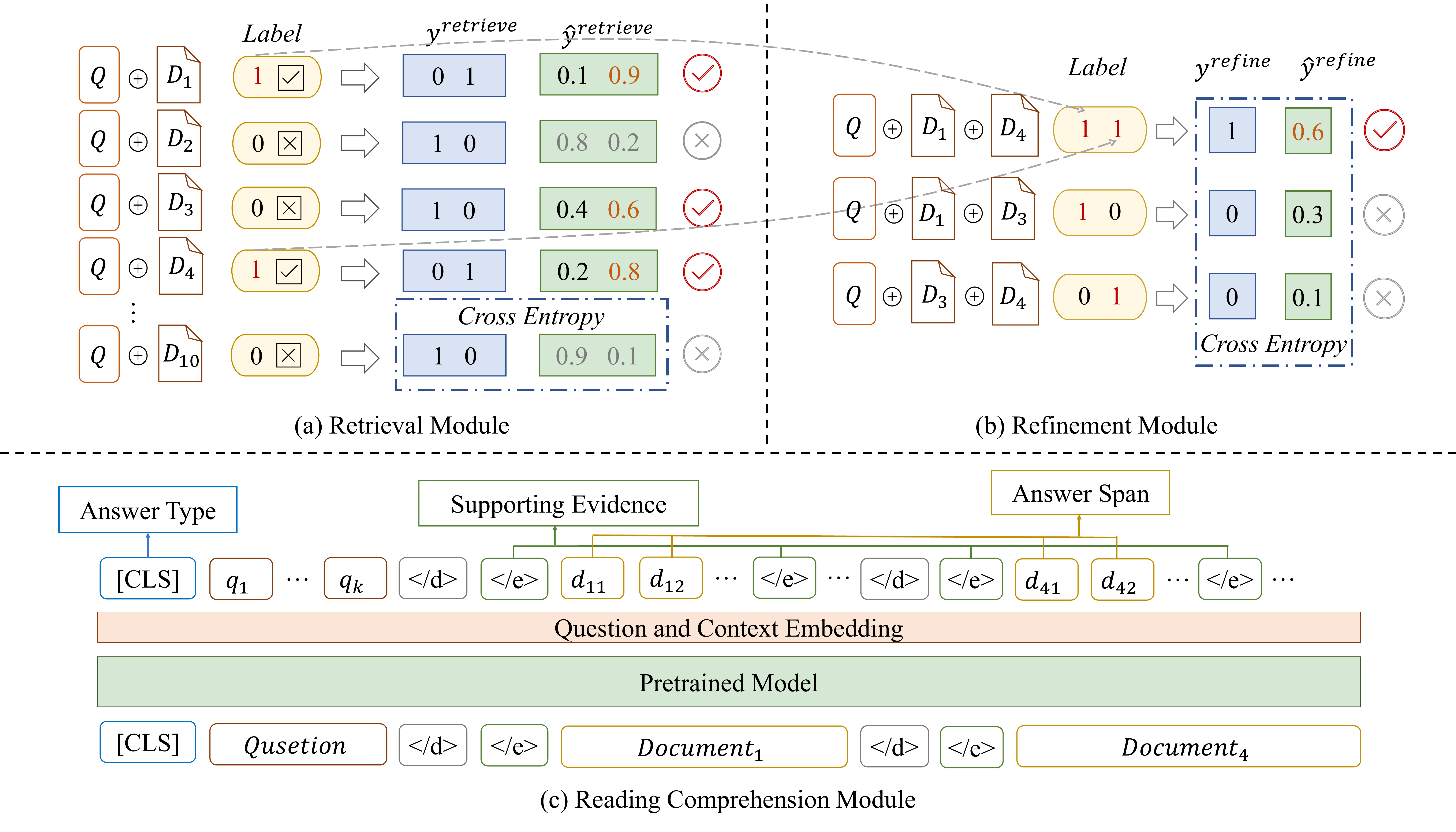}
    \caption{Overview of our $\mathbf{R}^3$ model, which consists of three main modules: \textbf{R}etrieval, \textbf{R}efinement, and \textbf{R}eading Comprehension. $\mathbf{R}^3$ sequentially executes each module to arrive at the final answer.}
    \label{fig:R3}
\end{figure*}

Figure~\ref{fig:R3} depicts the overall architecture of $\mathbf{R}^3$. The retrieval module sift through and exclude irrelevant documents, effectively selecting those that are pertinent to the question for utilization in the subsequent modules. In this example, Document 1, 3, and 4 are selected due to their higher relevance scores, while the other documents are filtered out. Subsequently, the refinement module further selects documents based on their combined relevance. In this case, Document 1 and 4 are chosen. Following this, the question and Document 1 and 4 are concatenated and used as input for the reading comprehension module. Within the reading comprehension module, we employ a multi-task approach to simultaneously train for supporting sentence prediction, answer span extraction, and answer type selection.

\subsection{Retrieval Module}
\label{sec:retrival_module}
In the retrieval module, each question $Q$ is typically accompanied by a set of $M$ documents ${D_1, D_2\dots,D_M}$, but only $C, |C| << M$ (two in HotpotQA) are labeled as relevant to the question $Q$. We model the retrieval process as a binary classification task. Specifically, for each question-document pair, we generate an input by concatenating ``[CLS], question, [SEP], document, [SEP]'' in sequence. We then feed the [CLS] token output from the model into a linear classifier. $\mathcal{L}_{\text{retrieve}}$ represents the cross-entropy between the predicted probability and the gold label. In contrast to S2G~\cite{wu2021graph}, which employs a complex pairwise learning-to-rank loss, we opt for a simple binary cross-entropy loss, as it maintains high performance while being significantly more efficient.

\begin{equation}
\label{eq:retrival_loss}
\begin{split}
    \mathcal{L}_{\text{retrieve}}= \mathbb{E}[-\frac{1}{M}\sum_{i=1}^{M}(y_{i}^{\text{retrieve}} \cdot log(\hat{y}_{i}^{\text{retrieve}}) 
    +(1-y_{i}^{\text{retrieve}}) \cdot log(1-\hat{y}_{i}^{\text{retrieve}}))],
\end{split}
\end{equation}
where $\hat{y}_{i}^{\text{retrieve}}$ is the probability predicted by the model and $y_{i}^{\text{retrieve}}$ is the ground-truth label. $M$ is the number of provided documents. $\mathbb{E}$ means the expectation of all samples.

\begin{equation}
\label{eq:coarse_label}
    y_{i}^{\text{retrieve}}=\left\{
        \begin{array}{cc}
        1  & D_i \text{ is a gold document.}   \\
        0  & D_i \text{ is a non-gold document.}   \\
        \end{array} \right .
\end{equation}

\subsection{Refinement Module}
\label{sec:refinement_module}
The refinement module is designed to identify document combinations that are capable of supporting the entire multi-hop reasoning processes. We combine the $K$ documents obtained from the retrieval module to form $C^2_K$ document pairs. These are concatenated into the following sequence: ``[CLS], question, [SEP], document1, [SEP], document2, [SEP]''. Similar to the retrieval module, we extract the [CLS] token output from the model and pass it through a classifier. Document pairs containing two gold documents are labeled as 1, while others are labeled as 0. We model multi-hop reasoning as a selection task, focusing on choosing document combinations that effectively convey complete multi-hop reasoning information to the subsequent modules.

\begin{equation}
\label{eq:fine_loss}
    \mathcal{L}_{\text{refine}}=\mathbb{E}[-\sum_{i=1}^{C^2_K}y_{i}^{\text{refine}}log(\hat{y}_{i}^{\text{refine}})],
\end{equation}
where $\hat{y}_{i}^{\text{refine}}$ is predicted document pair probability and $y_{i}^{\text{refine}}$ is the 
ground-truth label, $C^2_K$ is number of all combination. 

\begin{equation}
\label{eq:fine_label}
   y_{i}^{\text{refine}}=\left\{
        \begin{array}{cc}
          1  & \mathcal{C}_i \text{ consists of two gold documents.}  \\
          0  &  \text{otherwise.}
        \end{array} \right.
\end{equation}

We use a single pretrained language model as the encoder for both the retrieval and refinement module, and the final loss is a weighted sum of $\mathcal{L}_{\text{retrieve}}$ and $\mathcal{L}_{\text{refine}}$. $\lambda_1$ and $\lambda_2$ are accordingly coefficients of $\mathcal{L}_{\text{retrieve}}$ and $\mathcal{L}_{\text{refine}}$ . 

\begin{equation}
\label{eq:retrieval_loss}
    \mathcal{L}_{\text{total}}= \lambda_1 \mathcal{L}_{\text{retrieve}}+\lambda_2 \mathcal{L}_{\text{refine}}.
\end{equation}

\subsection{Reading Comprehension Module}
\label{sec:mrc_module}
In the reading comprehension module, we use multi-task learning to simultaneously predict supporting sentences and extract answer span. HotpotQA~\cite{Yang2018HotpotQAAD} contains samples labeled as ``yes'' or ``no''. The practice of splicing ``yes'' or ``no'' tokens at the beginning of the sequence~\cite{li2022easy} could corrupt the original text's semantic information. To avoid the impact of irrelevant information, we introduce an answer type selection header trained with a cross-entropy loss function.

\begin{equation}
\label{eq:answer_type_loss}
    \mathcal{L}_{\text{type}}=\mathbb{E}[-\sum_{i=1}^{3}y_{i}^{\text{type}}log(\hat{y}_{i}^{\text{type}})],
\end{equation}
where $\hat{y}_{i}^{\text{type}}$ represents the predicted probability of answer type generated by our model, and $y_{i}^{\text{type}}$ denotes the ground-truth label. Answer type includes ``yes'', ``no'' and ``span''. 

\begin{equation}
\label{eq:answer_type_label}
    y_{i}^{\text{type}}=\left\{
        \begin{array}{cc}
        0  & \text{Answer is no.}   \\
        1  & \text{Answer is yes.} \\
        2  & \text{Answer is a span.} \\
        \end{array} \right .
\end{equation}

To extract the span of answers, we use a linear layer on the contextual representation to identify the start and end positions of answers, and adopts cross-entropy as the loss function. The corresponding loss terms are denoted as $\mathcal{L}_{\text{start}}$ and $\mathcal{L}_{\text{end}}$ respectively. Similar to previous work S2G~\cite{wu2021graph} and FE2H~\cite{li2022easy}, we also inject a special placeholder token $</e>$ and use a linear binary classifier on the output of $</e>$ to determine whether a sentence is a supporting fact. The classification loss of the supporting facts is denoted as $\mathcal{L}_{\text{sup}}$, and we jointly optimize all of these objectives in our model.
\begin{equation}
\label{eq:reading_loss}
    \mathcal{L}_{\text{reading}}= \lambda_3 \mathcal{L}_{\text{type}}+\lambda_4 (\mathcal{L}_{\text{start}}+\mathcal{L}_{\text{end}})+\lambda_5\mathcal{L}_{\text{sup}}.
\end{equation}

\section{Label Smoothing}
\label{sec:label_smoothing}
Label smoothing is a regularization technique that aims to improve generalization in a classifier by modifying the ground truth labels of the training data. In the one-hot setting, the probability of the correct category $q(y|x)$ for a training sample $(x,y)$ is typically defined as 1, while the probabilities of all other categories $q(\urcorner y|x)$ are defined as 0. The cross-entropy loss function used in this setting is typically defined as follows:

\begin{equation} 
\label{eq:label_smoothing_distribuction}
\mathcal{L}=-\sum_{k=1}^{K}q(k|x)\log(p(k|x)),
\end{equation}
where $p(k|x)$ is the probability of the model's prediction for the $k$-th class. Specifically, label smoothing mixes $q(k|x)$ with a uniform distribution $u(k)$, independent of the training samples, to produce a new distribution $q'(k|x)$.
\begin{equation} \label{eq:label_smoothing}
q'(k|x)=(1-\epsilon)q(k|x)+\epsilon u(k),
\end{equation}
where $\epsilon$ is the weight controls the importance of $q(k|x)$ and $u(k)$ in the resulting distribution. $u(k)$ is construed as $\frac{1}{K}$ of the uniform distribution, where $K$ is the total number of categories. Next, we introduce two novel label smoothing methods.

\begin{algorithm}[t]
  \caption{Linear Decay Label Smoothing.}
  \label{alg:LDLA}
  \begin{algorithmic}[1]
  \REQUIRE Training epochs $n > 0$; Smoothing weight $\epsilon \in [0,1]$; Decay rate $\tau \in [0,1]$; Uniform distribution $u$
  \STATE \textbf{Initialize}: Model parameter $w_0 \in \mathcal{W}$;
  \STATE \textbf{Input}: Optimization algorithm $\mathcal{A}$
  \FOR{$i=0,1,\dots,n$}
  \STATE $\epsilon_i \gets \epsilon - i\tau$
  \IF{$\epsilon_i < 0$}
  \STATE $\epsilon_i \gets 0$
  \ENDIF
  \STATE sample$(x_t,y_t)$
  \STATE $y_t^{LS} \gets (1-\epsilon_i)y_i + \epsilon u$
  \STATE $w_{i+1} \gets \mathcal{A-}step(w_i;x_i,y_i^{LS})$
  \ENDFOR
  \end{algorithmic}
\end{algorithm}

\subsection{Linear Decay Label Smoothing}
\label{sec:LDLA}
Our proposed Linear Decay Label Smoothing Algorithm (LDLA) addresses the abrupt changes in training distribution caused by the two-stage approach of TSLA, which can negatively impact the training process. Compared to TSLA, LDLA progressively decays the smoothing weight at a constant rate per epoch, which facilitates a more gradual learning process.

Given a total of $n$ epochs in the training process and a decay size of $\tau$, the smoothing weight $\epsilon$ for the $i$-th epoch can be calculated as follows:

\begin{equation}
\label{eq:ldla}
    \epsilon_{i}=\left\{
        \begin{array}{cc}
        \epsilon - i\tau  & \epsilon - i\tau \ge 0   \\
        0  & \epsilon - i\tau < 0 \\
        \end{array} \right .
\end{equation}

Algorithm~\ref{alg:LDLA} provides a detailed overview of the LDLA algorithm. LDLA employs the concept of curriculum learning by gradually transitioning the model's learning target from a smoothed distribution to the original distribution throughout the training process. This approach methodically reduces uncertainty during training, enabling the model to progressively concentrate on more challenging samples. The gradual shift from learning under conditions of uncertainty to a state of certainty ensures the stability of the learning process. As a result, the LDLA algorithm facilitates a learning process that is more stable and efficient.

\subsection{F1 Smoothing}
\label{sec:F1_smoothing}
\begin{figure*}[t]
    \centering
 \includegraphics[width=1\linewidth]{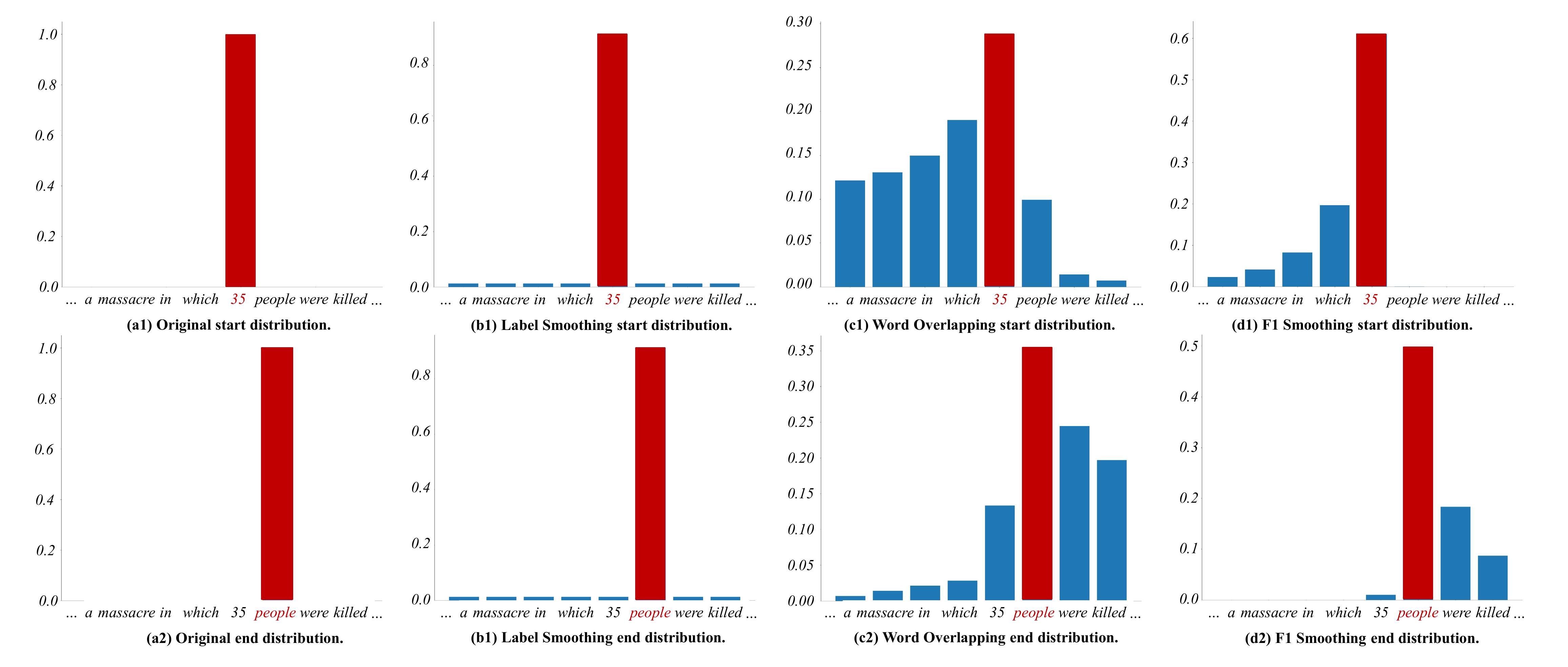}
    \vspace{-2em}
    \caption{
    Visualization of original distribution and different label smoothing distributions, including Label Smoothing, Word Overlapping, and F1 Smoothing. The first row shows the distribution of the start token, and the second row shows the distribution of the end token. The gold start and end tokens are highlighted in red. }
    \vspace{-.75em}
    \label{fig:distribution}
\end{figure*}

Unlike traditional classification tasks, MRC requires identifying both the start and end positions of a span. To address the specific nature of this task, a specialized smoothing method is required to prevent overfitting the specific answer span distribution in the training set. In this section, we introduce F1 Smoothing, a technique that calculates the significance of answer span based on its F1 score.

Consider a sample $x$ that contains a context $S$ and an answer $a_{gold}$. The total length of the context is denoted by $L$. We use $q_s(t|x)$ to denote the F1 score between a span of arbitrary length starting at position $t$ in $S$ and the ground truth answer $a_{\text{gold}}$. Similarly, $q_e(t|x)$ denotes the F1 score between $a_{\text{gold}}$ and a span of arbitrary length ending at position $t$ in $S$ .

\begin{equation}
\label{eq:start}
    q_s(t|x)=\sum_{\xi=t}^{L-1} F1\left((t,\xi),a_{\text{gold}}\right).
\end{equation}
\begin{equation}
\label{eq:end}
    q_e(t|x)=\sum_{\xi=0}^{t} F1\left((\xi,t),a_{\text{gold}}\right).
\end{equation}
The normalized distributions are noted as $q^{'}_s(t|x)$ and $q^{'}_e(t|x)$, respectively.
\begin{equation}
        q_s^{'}(t|x)=\frac{exp(q_s(t|x))}{\sum_{i=0}^{L-1}exp(q_s(i|x))}.
\end{equation}
\begin{equation}
        q_e^{'}(t|x)=\frac{exp(q_e(t|x))}{\sum_{i=0}^{L-1}exp(q_e(i|x))}.
\end{equation}

To decrease the computational complexity of F1 Smoothing, we present a computationally efficient version in Appendix~\ref{app:F1_optimization}. Previous research~\cite{zhao2020robust} has investigated various label smoothing methods for MRC, encompassing traditional label smoothing and word overlap smoothing. As illustrated in Figure~\ref{fig:distribution}, F1 Smoothing offers a more accurate distribution of token importance in comparison to Word Overlap method. This method reduces the probability of irrelevant tokens and prevents the model from being misled during training.

\section{Experiment}
\label{sec:Experiment}
\begin{table*}[t]
\centering
\begin{tabular}{lccccc}
\hline
\multirow{2}{*}{\textbf{Model}} & \multicolumn{2}{c}{\textbf{Answer}} & \multicolumn{2}{c}{\textbf{Supporting}} \\ \cline{2-5} 
                             & \textbf{EM}           & \textbf{F1}          & \textbf{EM}                 & \textbf{F1}                 \\ \hline
Baseline Model~\cite{Yang2018HotpotQAAD}               & 45.60        & 59.02       & 20.32              & 64.49              \\
QFE~\cite{nishida2019answering}                          & 53.86        & 68.06       & 57.75              & 84.49              \\
DFGN~\cite{qiu2019dynamically}                         & 56.31        & 69.69       & 51.50              & 81.62              \\
SAE-large~\cite{tu2020select}                    & 66.92        & 79.62       & 61.53              & 86.86              \\
C2F Reader~\cite{shao2020graph}                   & 67.98        & 81.24       & 60.81              & 87.63              \\
HGN-large~\cite{fang2019hierarchical}                    & 69.22        & 82.19       & 62.76              & 88.47              \\
FE2H on ELECTRA~\cite{li2022easy}              & 69.54        & 82.69       & 64.78              & 88.71              \\
AMGN+~\cite{li2021asynchronous}                        & 70.53        & 83.37       & 63.57              & 88.83              \\
S2G+EGA~\cite{wu2021graph}                      & 70.92        & 83.44       & 63.86              & 88.68              \\
FE2H on ALBERT~\cite{li2022easy}               & 71.89        & \textbf{84.44}       & 64.98              & 89.14              \\ \hline
$\mathbf{R}^3$ (ours)                   & 71.27        & 83.57       & 65.25              & 88.98              \\
Smoothing $\mathbf{R}^3$ (ours) & \textbf{72.07}        & 84.34       & \textbf{65.44}              & \textbf{89.55}              \\ \hline
\end{tabular}
\caption{In the distractor setting of the HotpotQA test set, our proposed F1 Smoothing and LDLA has led to significant improvements in the performance of the Smoothing $\mathbf{R}^3$ model compared to the $\mathbf{R}^3$ model. Furthermore, the Smoothing $\mathbf{R}^3$ model has outperformed a series of strong baselines and achieved remarkable state-of-the-art performance.}
\label{tab:test_result}
\end{table*}

\subsection{Dataset}
\label{sec:dataset}
We evaluate our approach on the distractor setting of HotpotQA~\cite{Yang2018HotpotQAAD}, a multi-hop question-answer dataset with 90k training samples, 7.4k validation samples, and 7.4k test samples. Each question in this dataset is provided with several candidate documents, two of which are labeled as gold. In addition, HotpotQA also provides supporting sentences for each question, encouraging the model to explain the reasoning path of the multi-hop question answering. We use the Exact Match (EM) and F1 score (F1) to evaluate the performance of our approach in terms of document retrieval, supporting sentence prediction, and answer extraction.

\subsection{Implementation Details} 
\label{sec:implementation_details}
Our model is built using the Pre-trained Language Models (PLMs) provided by HuggingFace's Transformers library~\cite{wolf2020transformers}.

\vspace{-.1em}
\paragraph{Retrieval and Refinement Module.} We use RoBERTa-large~\cite{liu2019roberta} and ELECTRA-large~\cite{clark2020electra} as our PLMs and conduct an analysis on RoBERTa-large~\cite{liu2019roberta} in Section~\ref{sec:analysis}. Training on a single RTX3090 GPU, we set the number of epochs to 8 and the batch size to 16. We employ the AdamW~\cite{loshchilov2017decoupled} optimizer with a learning rate of 5e-6 and a weight decay of 1e-2.

\vspace{-.1em}
\paragraph{Reading Comprehension Module.} We employ RoBERTa-large~\cite{liu2019roberta} and DeBERTa-v2-xxlarge~\cite{He2021DeBERTaV3ID} as our Pre-trained Language Models (PLMs), with our analyses primarily conducted using RoBERTa-large~\cite{liu2019roberta}. To train RoBERTa-large, we use an RTX3090 GPU, setting the number of epochs to 16 and the batch size to 16. For the larger DeBERTa-v2-xxlarge model, we employ an A100 GPU, setting the number of epochs to 8 and the batch size to 16. We use the AdamW optimizer~\cite{loshchilov2017decoupled} with a learning rate of 4e-6 for RoBERTa-large and 2e-6 for DeBERTa-v2-xxlarge, along with a weight decay of 1e-2 for optimization.

\begin{table}[t]
\centering
\begin{tabular}{lccccl}
\hline
\textbf{Model} & \textbf{EM} & \textbf{F1} \\
\hline
 $\text{SAE}_{large}$~\cite{tu2020select}& 91.98 & 95.76 \\
 $\text{S2G}_{large}$~\cite{wu2021graph}& 95.77 & 97.82 \\
$\text{FE2H}_{large}$~\cite{li2022easy}& 96.32 & 98.02 \\
\hline
$\mathbf{R}^3$ (ours)& 96.50 & 98.10 \\
Smoothing $\mathbf{R}^3$ & \textbf{96.85} & \textbf{98.32} \\
\hline
\end{tabular}
\caption{\label{tab:RE_results}
Comparison of our $\mathbf{R}^3$ and Smoothing $\mathbf{R}^3$ model with several strong baselines in document retrieval task on HotpotQA validation set. Smoothing $\mathbf{R}^3$ model demonstrates further performance enhancement compared to $\mathbf{R}^3$.}
\vspace{-1em}
\end{table}
\begin{figure*}[t]
    \centering
    \begin{minipage}{0.48\textwidth}
    \begin{minipage}{0.48\textwidth}
    \centering
    \includegraphics[width=1\linewidth]{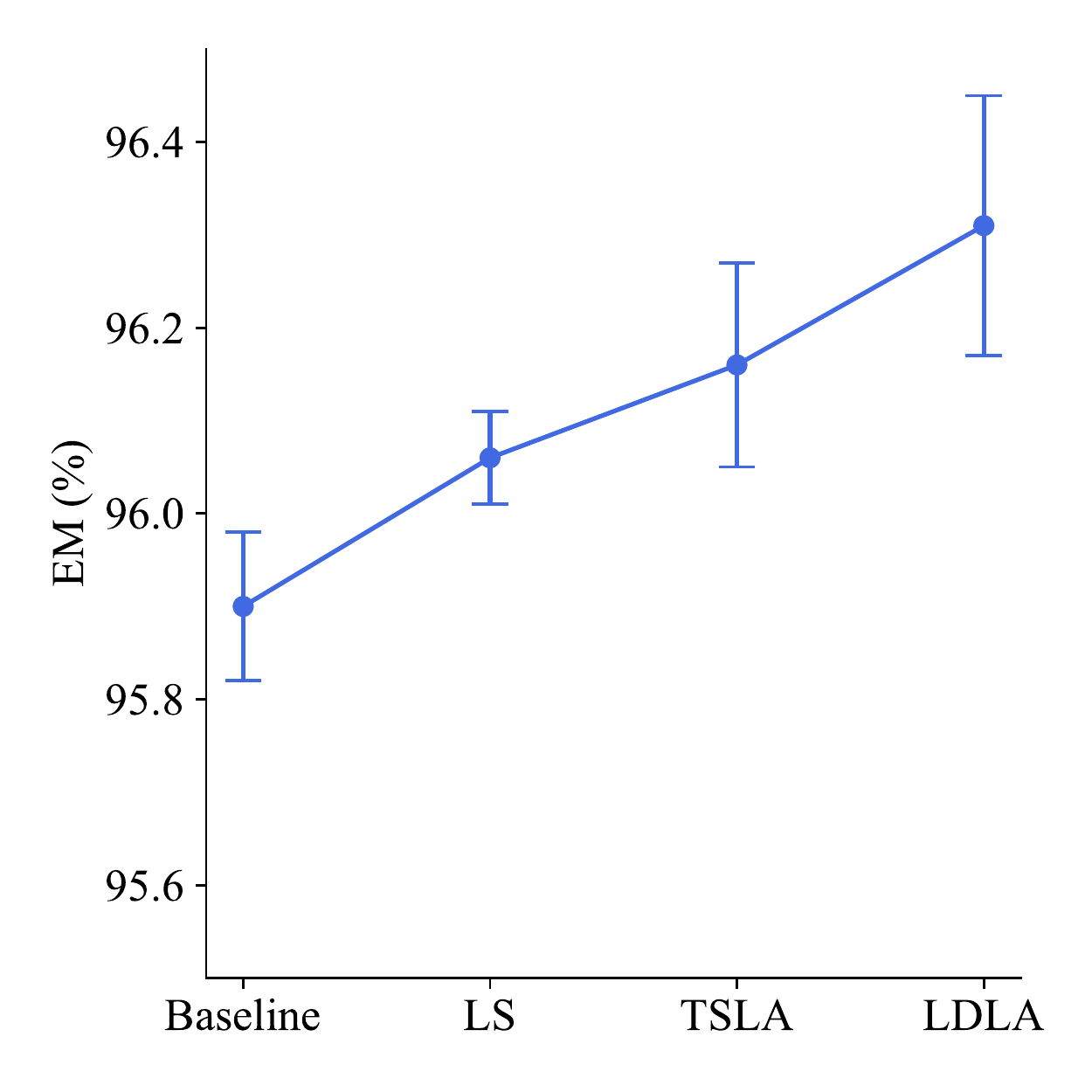}
    \end{minipage}
    \hfill
    \begin{minipage}{0.48\textwidth}
    \centering
    \includegraphics[width=1\linewidth]{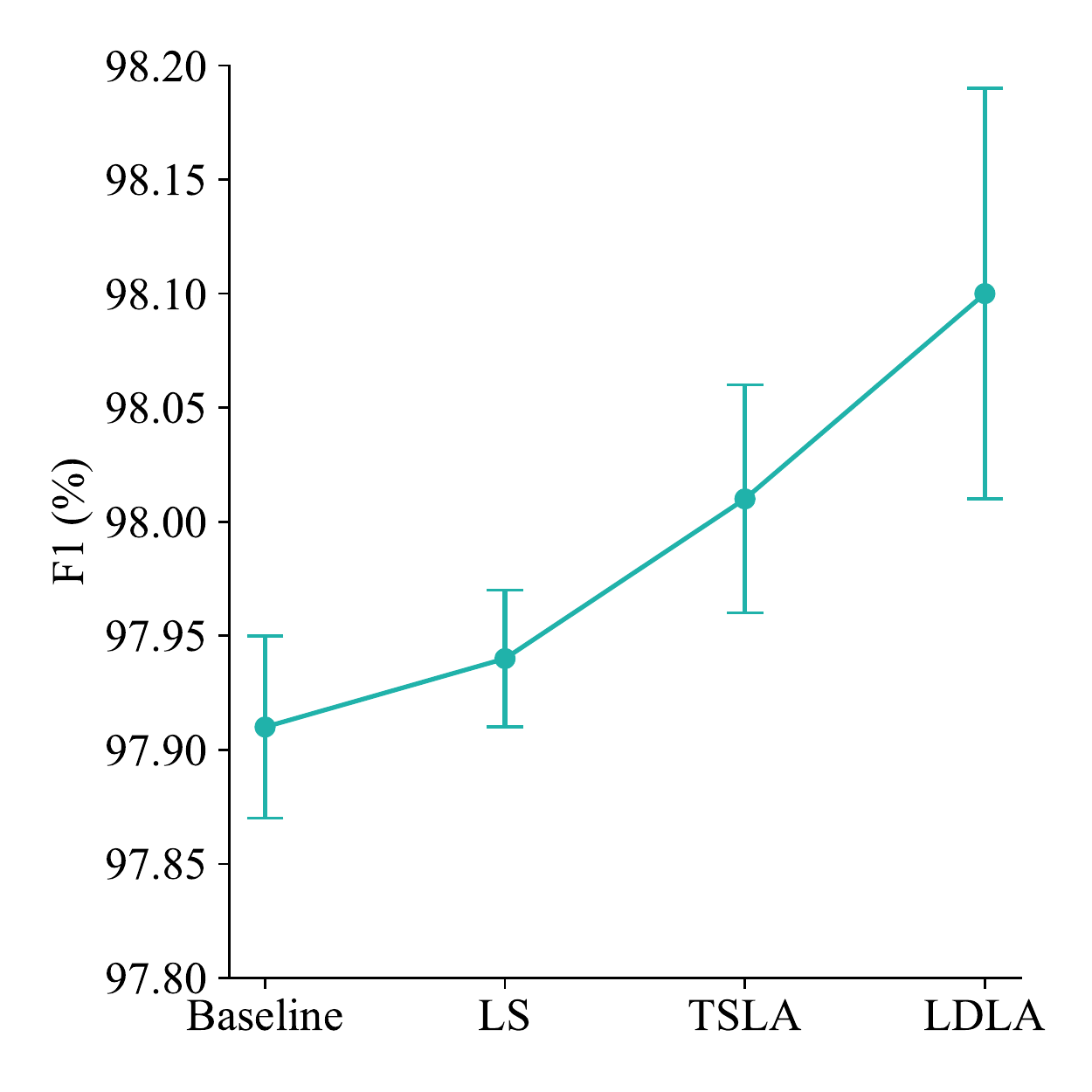}
    \end{minipage}
    \caption{Performance comparison of a series of smoothing methods in document retrieval task.}
    \label{fig:exp-document-retrieval}
    \end{minipage}
    \hfill
    \begin{minipage}{0.48\textwidth}
    \begin{minipage}{0.48\textwidth}
    \centering
    \includegraphics[width=1\linewidth]{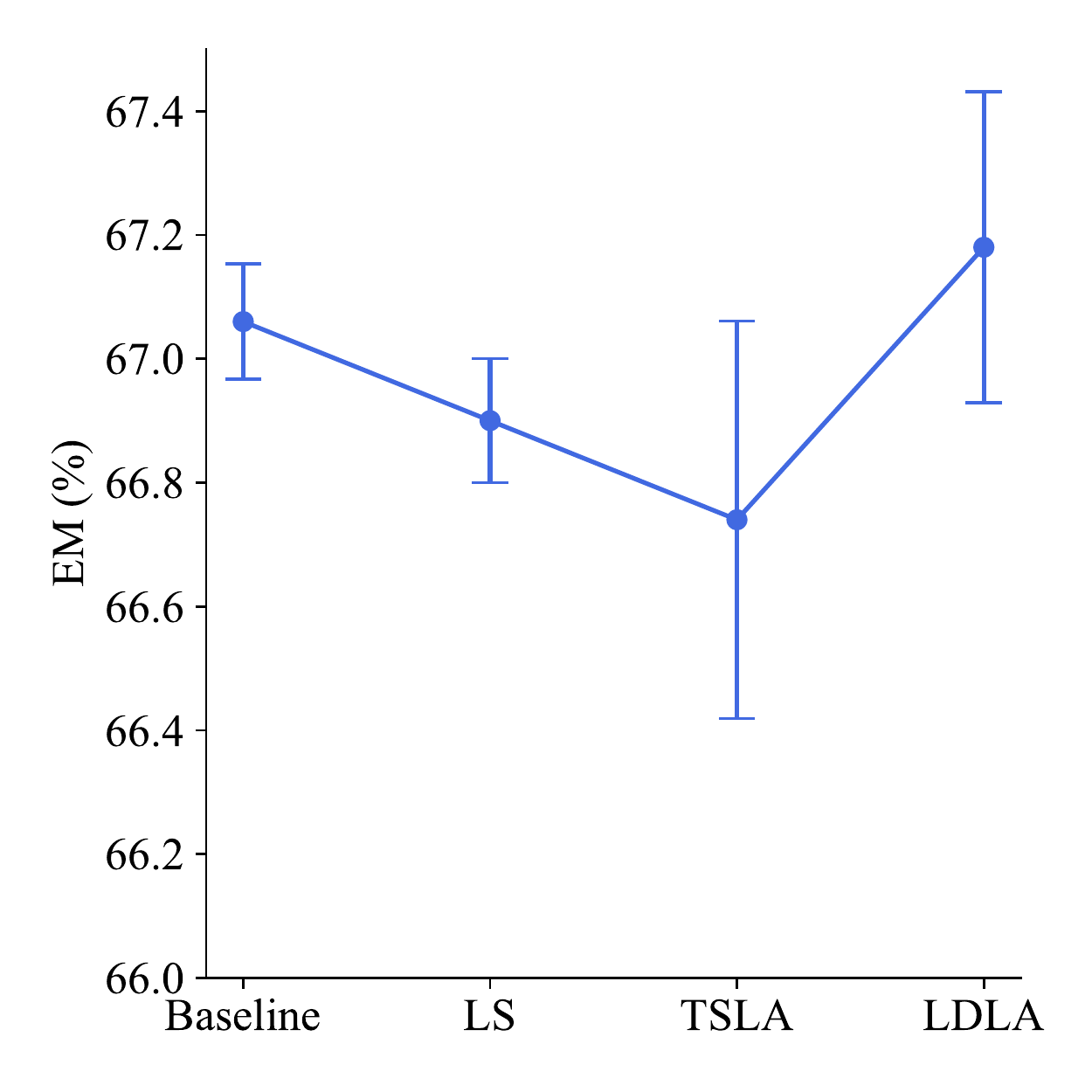}
    \end{minipage}
    \hfill
    \begin{minipage}{0.48\textwidth}
    \centering
    \includegraphics[width=1\linewidth]{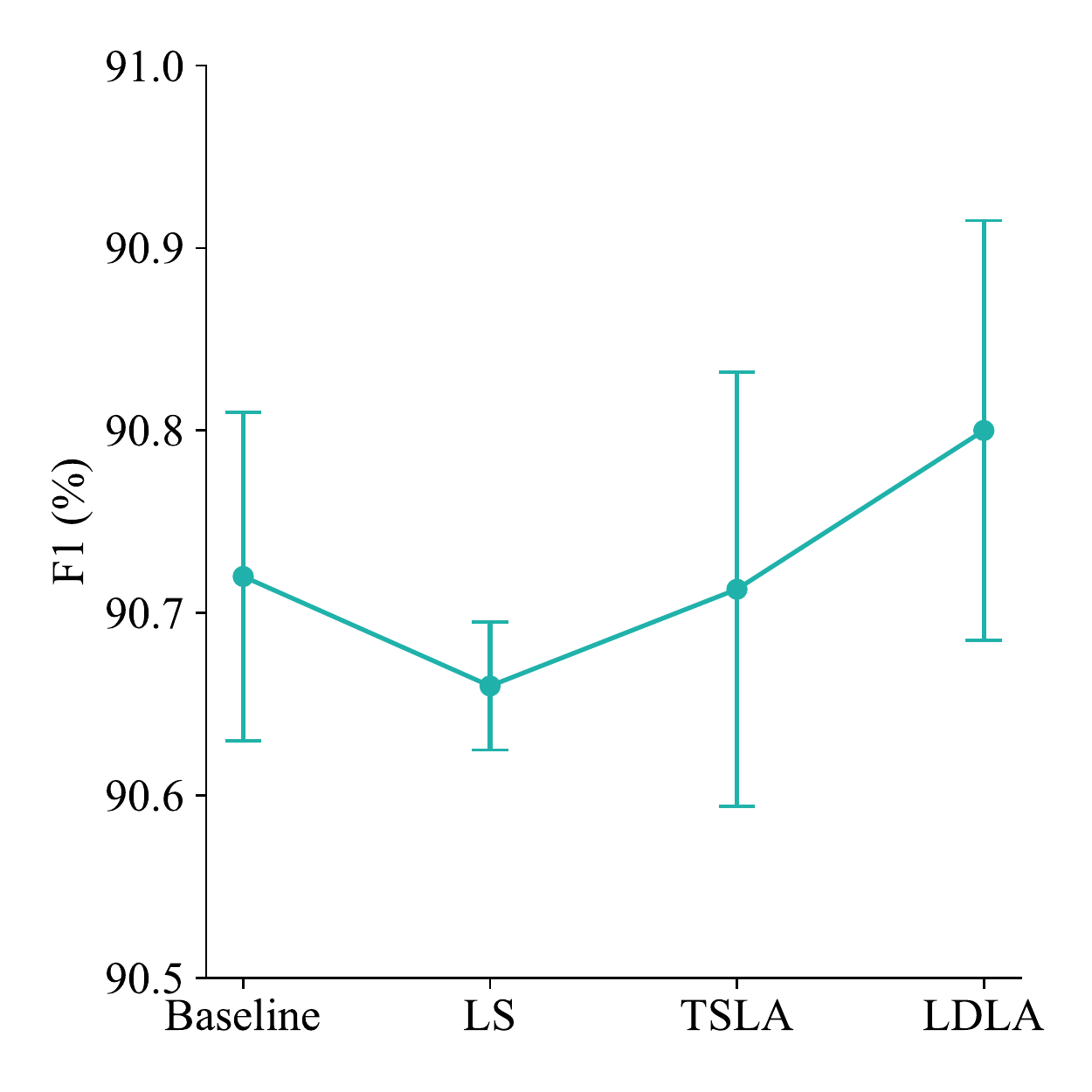}
    \end{minipage}
    \caption{Comparison of various smoothing methods in supporting sentence prediction task.}
    \label{fig:exp-supporting-sentence-prediction}
    \end{minipage}
    \label{fig:example}
\end{figure*}

\begin{table}[t]
\centering
\begin{tabular}{ccccc}
\hline
\multirow{2}{*}{\textbf{Model}}       & \multicolumn{2}{c}{\textbf{Answer}} & \multicolumn{2}{c}{\textbf{Supporting}} \\ \cline{2-5} 
                             & \textbf{EM}           & \textbf{F1}          & \textbf{EM}                 & \textbf{F1}                 \\ \hline
SAE                          & 67.70        & 80.75       & 63.30              & 87.38              \\
S2G                          & 70.80        & -           & 65.70              & -                  \\ \hline
$\mathbf{R}^3$                   & 71.39        & 83.84       & 66.32              & 89.54              \\
Smoothing $\mathbf{R}^3$ & \textbf{71.89}        & \textbf{84.65}       & \textbf{66.75}              & \textbf{90.08}              \\ \hline
\end{tabular}
\caption{Comparison of cascade results between our method and several previous methods on the validation set of HotpotQA.}
\label{tab:cascade_results}
\vspace{-1em}
\end{table}

\subsection{Experimental Results} 
\label{sec:experiment_results}
We utilize ELECTRA-large~\cite{clark2020electra} as the PLM for the retrieval and refinement modules, and DeBERTa-v2-xxlarge for the reading comprehension module. The $\mathbf{R}^3$ model incorporating F1 Smoothing and LDLA methods is referred to as Smoothing $\mathbf{R}^3$. LDLA is employed for document retrieval and supporting sentence prediction, while F1 Smoothing is applied for answer span extraction. As shown in Table~\ref{tab:test_result}, compared to a series of previous strong baselines, Smoothing $\mathbf{R}^3$ has achieved the best performance on the stringent Exact Match (EM) metric. Additionally, compared to $\mathbf{R}^3$, Smoothing $\mathbf{R}^3$ shows an improvement of 0.8\% and 0.77\% in EM and F1 scores for the answer extraction task. For the supporting sentence prediction task, there is an increase of 0.19\% and 0.57\% in EM and F1 scores. These results indicate that label smoothing effectively enhances the model's performance across various metrics.

\vspace{-.1em}
\paragraph{Document Retrieval.} We compare the performance of our retrieval and refinement module, using ELECTRA-large as a backbone, to three advanced methods: SAE~\cite{tu2020select}, S2G~\cite{wu2021graph}, and FE2H~\cite{li2022easy}. These methods employ sophisticated selectors for retrieving relevant documents. We evaluate the performance of document retrieval using the EM and F1 metrics. Table~\ref{tab:RE_results} demonstrates that our $\mathbf{R}^3$ method outperforms these three strong baselines, with Smoothing $\mathbf{R}^3$ further enhancing performance.

\vspace{-.1em}
\paragraph{Supporting Sentence Prediction and Answer Span Extraction.} In Table~\ref{tab:cascade_results}, we evaluate the performance of the reading comprehension module, which employs DeBERTa-v2-xxlarge~\cite{He2021DeBERTaV3ID} as the backbone, on documents retrieved by the retrieval and refinement module. Our $\mathbf{R}^3$ model outperforms strong baselines SAE and S2G, and further improvements are achieved by incorporating F1 Smoothing and LDLA. These results emphasize the potential for enhancing performance through the application of label smoothing techniques.

\subsection{Label Smoothing Analysis}
\label{sec:analysis}
In our analysis of label smoothing, we use RoBERTa-large~\cite{liu2019roberta} as the backbone. To ensure the reliability of our experimental results, we conduct multiple runs with different random number seeds (41, 42, 43, and 44).

In our experiments, we compare three label smoothing methods: Label Smoothing (LS), Two-Stage Label smoothing (TSLA), and Linear Decay Label smoothing (LDLA). The initial value of $\epsilon$ in our experiments was 0.1, and in the first stage of TSLA, the number of epochs was set to 4. For each epoch in LDLA, $\epsilon$ was decreased by 0.01.

\vspace{-.1em}
\paragraph{Document Retrieval.} As shown in Figure~\ref{fig:exp-document-retrieval}, label smoothing effectively enhances the generalization performance of the retrieval module. LDLA label smoothing approach has more effectively enhanced the model's performance in document retrieval tasks compared to other label smoothing methods.

\vspace{-.1em}
\paragraph{Supporting Sentence Prediction.} We assess the impact of label smoothing on the supporting sentence prediction task. As illustrated in Figure~\ref{fig:exp-supporting-sentence-prediction}, we observe that label smoothing and the TSLA method do not exhibit significant advantages over the baseline and even lead to decreased performance. In contrast, our proposed LDLA method effectively improves the model's performance in the sentence prediction task. This demonstrates the broader task applicability and effectiveness of the LDLA method.

\begin{figure*}[t]
    \centering
    \begin{minipage}{0.48\textwidth}
    \centering
    \includegraphics[width=1\linewidth]{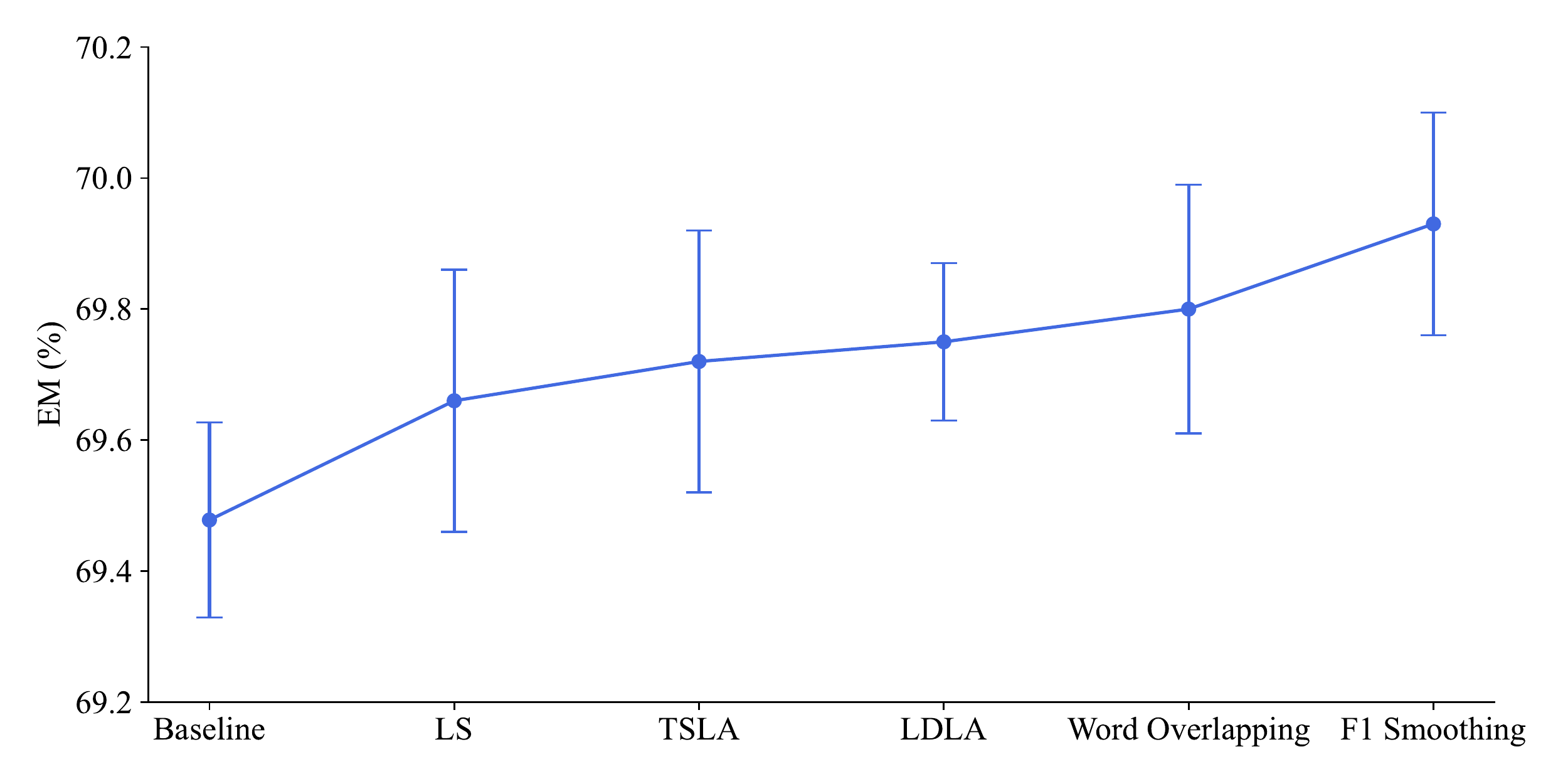}
    \end{minipage}
    \hfill
    \begin{minipage}{0.48\textwidth}
    \centering
    \includegraphics[width=1\linewidth]{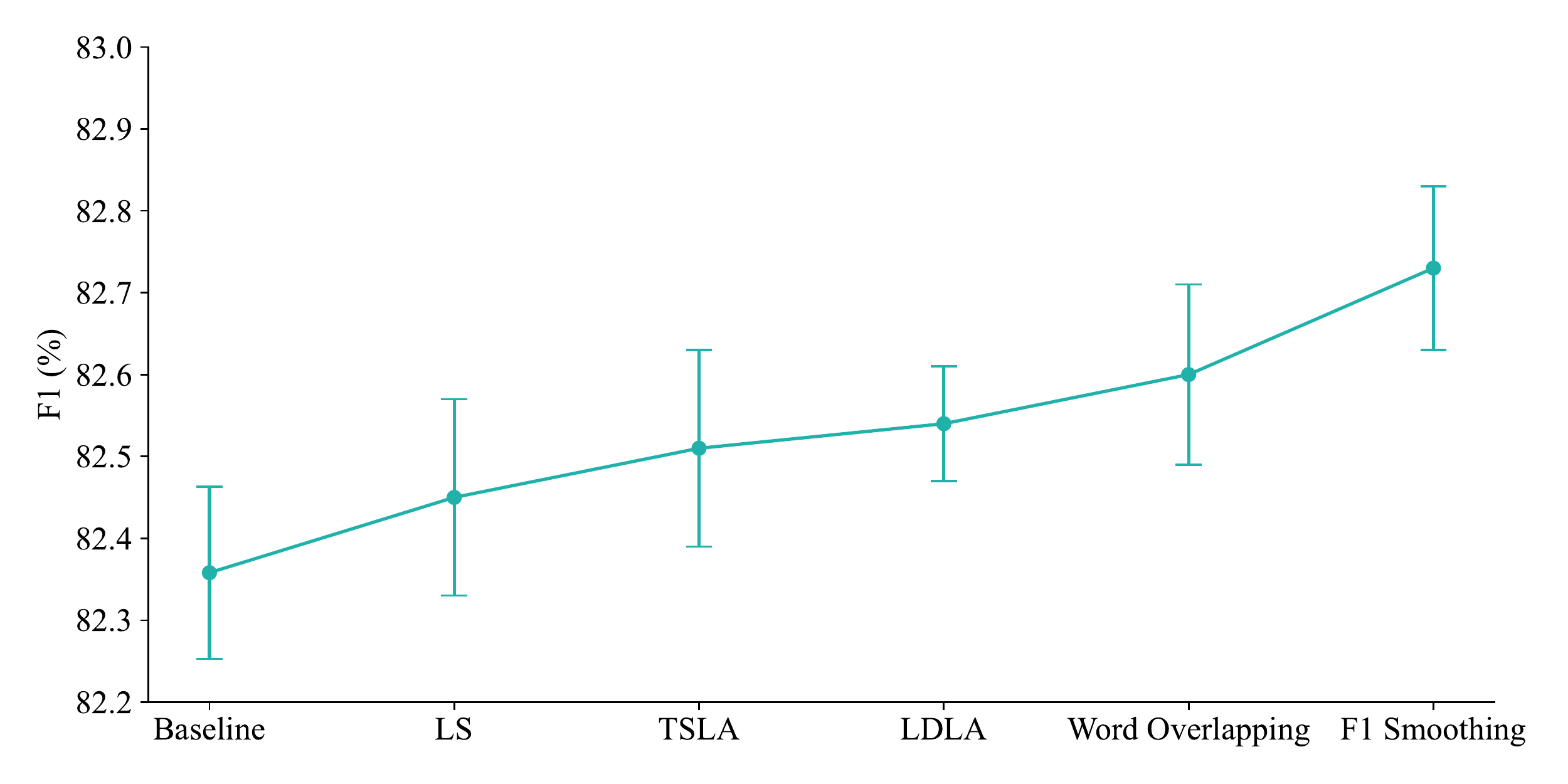}
    \end{minipage}
    \caption{Analysis of different label smoothing methods in answer span extraction task. }
    \label{fig:exp-answer-span-extraction}
\end{figure*}

\vspace{-.1em}
\paragraph{Answer Span Extraction.} The impact of label smoothing methods on answer span extraction in the reading comprehension module is depicted in Figure~\ref{fig:exp-answer-span-extraction}. Compared to the baseline, methods such as label smoothing, TSLA, and LDLA can mitigate the performance decline caused by overfitting, thereby enhancing the model's performance. F1 Smoothing shows a significant improvement over label smoothing, TSLA, and LDLA, and also has a notable advantage over the Word Overlapping method. This indicates that F1 Smoothing, by assigning appropriate weights to different tokens in a sentence, more effectively and precisely calculates the suitable target distribution, thereby significantly improving the model's performance in answer span extraction tasks.

\subsection{Error Analysis}
\label{sec:error_analysis}
\begin{table}[t]
\centering
\begin{tabular}{cccc}
\hline
\textbf{Model}         & \textbf{Answer Span Errors} & \textbf{Multi-Hop Reasoning Errors} \\ \hline
S2G                   & 1612                  & 550                   \\
$\mathbf{R}^3$        & 1556                  & 562                   \\
Smoothing $\mathbf{R}^3$  & 1536 ($\downarrow 1.3\%$)       & 545($\downarrow 3.0\%$)                  \\ \hline
\end{tabular}
\caption{\label{tab:error_analysis} Error analysis on Answer Span Errors and Multi-hop Reasoning Errors.}
\vspace{-1em}
\end{table}

To more comprehensively understand the role of label smoothing in enhancing model performance, our analysis delves into the model's outputs on the validation set, with a particular emphasis on answer span and multi-hop reasoning errors. These errors are defined as follows:

\begin{itemize}
\item Answer Span Errors: Occur when the model's predicted answer and the ground truth answer share some overlap (after excluding stop words) but are not entirely the same.
\item Multi-hop Reasoning Errors: Arise when the model's reasoning process leads to a predicted answer that is entirely different from the ground truth answer.
\end{itemize}

The implementation of label smoothing has led to notable improvements, as detailed in Table~\ref{tab:error_analysis}. Specifically, Smoothing $\mathbf{R}^3$ achieved a 1.3\% reduction in answer span errors, decreasing from 1556 to 1536 instances, and a 3.0\% decrease in multi-hop reasoning errors, reducing the count from 562 to 545. These reductions in both error types are significant when compared to the performance of the S2G model. This evidence strongly suggests that label smoothing, when integrated during training, can prevent the model from excessively fitting to specific answer spans and reasoning pathways found in the training set. Consequently, this leads to enhanced generalization capabilities and improved overall performance of the model.

\section{Conclusion}
\label{sec:conclusion}
In this study, we first identify the primary challenges hindering the performance of MHQA systems and propose using label smoothing to mitigate overfitting issues during MHQA training. We introduce F1 smoothing, a novel smoothing method inspired by the widely-used F1 score in MRC tasks. Additionally, we present LDLA, a progressive label smoothing algorithm that incorporates the concept of curriculum learning. Comprehensive experiments on the HotpotQA dataset demonstrate that our proposed model, Smoothing $\mathbf{R}^3$, achieves significant performance improvement when using F1 smoothing and LDLA. Further analysis indicates that label smoothing is a valuable technique for MHQA, effectively improving the model's generalization while minimizing overfitting to particular patterns in the training set.

\section*{Acknowledgement}
\label{sec:acknowledge}
We would like to express our heartfelt thanks to the students and teachers of FudanNLP Lab. Their thoughtful suggestions, viewpoints, and enlightening discussions have made significant contributions to this work. We also greatly appreciate the strong support from Huawei Poisson Lab for our work, and their invaluable advice. We are sincerely grateful to the anonymous reviewers and the domain chairs, whose constructive feedback played a crucial role in enhancing the quality of our research. This work was supported by the National Key Research and Development Program of China (No.2022CSJGG0801), National Natural Science Foundation of China (No.62022027) and CAAI-Huawei MindSpore Open Fund.

\bibliographystyle{ccl}
\bibliography{ccl2022-en}

\section{Appendix A}

\subsection{Implementation Details}
\label{app:Implementation_details}
Our \href{https://github.com/yinzhangyue/Smoothing-R3}{Github} repository showcases more details and specific implementations.

\subsection{Efficient F1 Smoothing}
\label{app:F1_optimization}
In order to alleviate the complexity introduced by multiple for loops in the F1 Smoothing method, we have optimized Eq.~\eqref{eq:start} and Eq.~\eqref{eq:end}. We use $L_a=e^{*}-s^{*}+1$ and $L_p=e-s+1$ to denote respectively the length of gold answer and predicted answer.
\begin{equation}
    q_s(t|x)=\sum_{\xi=t}^{L-1} \text{F1}\left((t,\xi),a_{\text{gold}}\right).
\end{equation}
If $t < s^{*}$, the distribution is
\begin{equation}
\label{eq:qs1}
   q_s(t|x)= \sum_{\xi=s^{*}}^{e^{*}} \frac{2(\xi-s^{*}+1)}{L_p+L_a} + \sum_{\xi=e^{*}+1}^{L-1} \frac{2L_a}{L_p+L_a},
\end{equation}
else if $s^{*} \le t \le e^{*}$, we have the following distribution
\begin{equation}
\label{eq:qs2}
  q_s(t|x)=\sum_{\xi=s}^{e^{*}} \frac{2L_p}{L_p+L_a} + \sum_{\xi=e^{*}+1}^{L-1} \frac{2(e^{*}-s+1)}{L_p+L_a}.  
\end{equation}
In equation \ref{eq:qs1} and \ref{eq:qs2}, $L_p=e-i+1$.

 We can get $q_e(t|x)$ similarly. If $t > e^{*}$,
\begin{equation}
\label{eq:qe1}
    q_e(t|x)= \sum_{\xi=s^{*}}^{e^{*}} \frac{2(e^{*}-\xi+1)}{L_p+L_a} + \sum_{\xi=0}^{s^{*}-1} \frac{2L_a}{L_p+L_a},
\end{equation}
else if $s^{*} \le t \le e^{*}$,
\begin{equation}
\label{eq:qe2}
    q_e(t|x)= \sum_{\xi=s^{*}}^{e} \frac{2L_p}{L_p+L_a} + \sum_{\xi=0}^{s^{*}-1} \frac{2(e-s^{*}+1)}{L_p+L_a}.
\end{equation}
In equation \ref{eq:qe1} and \ref{eq:qe2}, $L_p=i-s+1$.

\end{document}